\documentclass[%
 reprint,
 amsmath,amssymb,
 aps,
]{revtex4-2}

\usepackage{graphicx}
\usepackage{dcolumn}
\usepackage{bm}

\usepackage{algorithmic}
\usepackage{algorithm}
\usepackage{color}
\usepackage{amsmath}
\usepackage{graphicx}
\usepackage{caption}
\usepackage{amsmath}
\usepackage{subfigure}

\usepackage[justification=centering]{caption}

\usepackage{marvosym}

\begin{document}

\preprint{APS/123-QED}

\title{Accelerate Support Vector Clustering via
Spectrum-Preserving Data Compression}
\thanks{Yuxuan Song and Yongyu Wang are co-first authors and  contributed equally to this paper. Yongyu Wang is the corresponding author. Correspondence to wangyongyu1@jd.com}%

\author{Yuxuan Song*}
 
\author{Yongyu Wang\textsuperscript{\Letter}*}%
 
\affiliation{%
 JD Logistics\\
}%

\maketitle

This paper proposes a novel framework for accelerating support vector clustering. The proposed method first computes much smaller compressed data sets while preserving the key
cluster properties of the original data sets based on a novel spectral data compression approach. Then, the resultant spectrally-compressed data sets are leveraged for the development of fast and high quality algorithm for support vector clustering. We conducted extensive experiments using real-world data sets and obtained very promising results. The proposed method allows us to achieve 100X and 115X speedups over the state of the art SVC method on the Pendigits and USPS data sets, respectively, while achieving even better clustering quality. To the best of our knowledge, this represents the first practical method for high-quality and fast SVC on large-scale real-world data sets

\section{Introduction}
\label{sec:intro}

Support vector approach have played dominant role in machine learning field for two decades. To solve the supervised classification problem, support vector machine (SVM) constructs a hyperplane in a high-dimensional space via support vectors. Extended from SVM, \cite{ben2001support} proposed support vector clustering (SVC) method, employing support vectors to discover the boundaries of clusters. Compared to traditional clustering methods such as $k$-means, SVC can well-detect linearly inseparable clusters by mapping the samples into a high-dimensional feature space and does not require people to specify the number of clusters in advance, which is highly desired for unseen data sets. When comparing with the DBSCAN clustering method, SVC can well-detect clusters from any density scenario.

Due to the rigorous theoretical foundations and many desired characteristic, SVC has drawn great attention from both the research and industrial communities. However, the cluster assignment step of the original SVC has a \textbf{$O(N^2)$} time complexity which can immediately hinder its applications in emerging big data analytics tasks. Another main drawback of the original SVC is that it cannot handle boundary support vectors (BSVs). In large-scale real-world data sets where a lot of BSVs exist, only a small portion of points can be clustered by the original SVC method and a large amount of BSVs are mis-recognized as outliers.

To solve the above problems, substantial effort has been devoted and many methods have been developed. Among them, two representative works are most important: \cite{yang2002support} attempted to accelerate SVC through proximity graphs. However, it cannot solve the BSV problem. \cite{lee2005improved} proposed a stable equilibrium point (SEP)-based method that can solve the BSV problem very well. However, the SEP calculation process requires to perform gradient descent on every data point which is an extremely time-consuming process for large-scale data set. So as acknowledge in \cite{lee2005improved}, it is impractical to do it on thousands of data points. Alas, from then, there was almost no meaningful progress on SVC.

In this paper, we propose to fundamentally address the two issues of SVC simultaneously via a novel spectrum-preserving data compression method. Our method first computes a much smaller compressed data set while preserving the key cluster properties of the original data set, by exploiting a nearly-linear time spectral data compression approach. The compressed data set enables to dramatically reduce the size of search space for SEP searching process. Then, the resultant SEPs can be directly applied to the SEP-based SVC algorithm. Due to the guaranteed preservation of the original spectrum for clustering purpose, the proposed method can dramatically improve the efficiency of the SEP-based SVC without loss of solution quality. To the best of our knowledge, this represents the first practical method for high-quality and fast SVC on large-scale real-world data sets. Another distinguished  characteristic of the proposed method is that it allows users can set the compression ratio based on their needs. Experimental results show that with 2X, 5X, 10X reductions, the proposed method consistently produces satisfactory clustering quality on real-world data sets.

\section{Preliminary}

\subsection{Support Vector Clustering}

Given data set $D$ with $N$ samples $\mathbf{x}_1,...,\mathbf{x}_N$, the original SVC algorithm find clusters with two main steps, namely SVM training step and cluster labeling step. The SVM training step constructs a trained kernel radius function by first mapping the samples to high dimensional feature space with a nonlinear transformation $\Phi$, then finding the minimal hypersphere of radius $R$ with the following constraints:

\begin{equation}
{{\|\Phi(\mathbf{x}_j)-\mathbf{a}\|}}^{2} \leq R^{2}+\xi_j,
\end{equation}
where $\mathbf{a}$ denotes the center of the hypersphere and $\xi_j \geq 0$ denotes the slack variable. 



To form the dual problem for solving the problem (1), the following Lagrangian is introduced:

\begin{equation}
\begin{split}
L=R^{2}-\sum_{j}\left(R^{2}+\xi_{j}-\left\|\Phi\left(\mathbf{x}_{j}\right)-\mathbf{a}\right\|^{2}\right) \beta_{j} \\
-\sum \xi_{j} \mu_{j}+C \sum \xi_{j},
\end{split}
\end{equation}
where $\beta_j$ and $\mu_j$ are the Lagrangian multipliers, $C$ is a constant set manually with the constraints: $0 \leq \beta_j \leq C$. Then, the solution of (1) can be find by solving the following optimization problem:\\

\begin{equation}
\centerline{Maximize: $W= \displaystyle\sum_{j} \Phi(\mathbf{x}_j)^{2}\beta_j-\displaystyle\sum_{i,j} \beta_i \beta_j\Phi(\mathbf{x}_i)\Phi(\mathbf{x}_j)$}
\end{equation}

\centerline{Subject to: $0 \leq \beta_j \leq C, \displaystyle\sum_{j} \beta_j=1, j=1,...,N.$}

The points that satisfy $0 < \beta_j < C$ are called support vectors (SVs) and the boundaries of clusters are formed by the SVs. The points that satisfy $\beta_j = 0$ lie inside clusters and the points satisfy $\beta_j = C$ are called bounded support vectors (BSVs) which cannot be clustered by the original SVC method. 

By using the Mercer kernel ( $K(\mathbf{x}_i, \mathbf{x}_j) = \mathit{e}^{-q \left\| \mathbf{x}_i - \mathbf{x}_j \right\|^2}$) to replace the inner product of $\Phi(\mathbf{x}_i) \cdot \Phi(\mathbf{x}_j)$, the following trained kernel radius function can be obtained:

\begin{equation}
\begin{split}
f(\mathbf{x}) &= R^2(\mathbf{x}) = {{\|\Phi(\mathbf{x})-\mathbf{a}\|}}^{2}\\
&= K(\mathbf{x}, \mathbf{x}) - 2\sum_{j} K(\mathbf{x}_j, \mathbf{x}_j)\beta_j + \sum_{i,j} \beta_i \beta_j K(\mathbf{x}_i, \mathbf{x}_j)
\end{split}
\end{equation}

Then, in the cluster labeling stage, for each pair of points, SVC needs to check whether any segment point $\mathbf{y}$ between them satisfy $R(\mathbf{y}) < R$ , which is a very time consuming step.

\subsection{Graph Laplacian Matrices}
Consider a graph $G=(V,E,w)$, where $V$ is the vertex set of the graph, $E$ is the edge set of the graph, and $w$ is a weight function that assign positive weights to all edges. The Laplacian matrix of graph G is a symmetric diagonally dominant (SDD) matrix defined as follows:
\begin{equation}\label{formula_laplacian}
L_G(p,q)=\begin{cases}
-w(p,q) & \text{ if } (p,q)\in E \\
\sum\limits_{(p,t)\in E}w(p,t) & \text{ if } (p=q) \\
0 & \text{ if } otherwise.
\end{cases}
\end{equation}

According to spectral graph theory \cite{chung1997spectral}, the clustering structure is embedded in the spectrum of the graph Laplacian and spectral information is critical for clustering.

\section{Methods}

\subsection{Algorithmic Framework}
Given a set of data samples, we first construct a $k$-NN graph $G$ to capture the basic manifold of the data set. Then we can obtain its corresponding graph Laplacian matrix $L_G$ accordingly. The spectral similarity between two points can be measured with the following steps: 
\begin{itemize}
    \item [$\bullet$]Calculate the bottom $K$ nontrivial graph Laplacian eigenvectors;
\end{itemize}

\begin{itemize}
    \item [$\bullet$]Construct a matrix $U$ with the calculated $K$ eigenvectors stored as column vectors;
\end{itemize}

\begin{itemize}
    \item [$\bullet$]Use each row of $U$ as the embedded feature vector of a point in low-dimensional space;
\end{itemize}

\begin{itemize}
    \item [$\bullet$]The spectral similarity between two points $u$ and $v$ can be defined as:
\end{itemize}

\begin{equation}
s_{uv} := \frac{{|(\mathbf{X}_u, \mathbf{X}_v)|}^2}{(\mathbf{X}_u, \mathbf{X}_u)(\mathbf{X}_v, \mathbf{X}_v)}, \mbox{   } (\mathbf{X}_u,\mathbf{X}_v) := \sum_{k=1}^{K}{\left(\mathbf{x}_u^{(k)} \cdot \mathbf{x}_v^{(k)}\right)}.
\end{equation}
where $\mathbf{x}_u$ and $\mathbf{x}_v$ are the embedded feature vectors of point $u$ and point $v$, respectively. The statistical interpretation is that $s_{uv}$ is the fraction of explained variance of linearly regressing $x_u$ on $x_v$ from spectral perspective. A higher value of $s_{uv}$ indicates stronger spectral correlation between the two points. In practice, to avoid the computational expensive eigen-decomposition procedure, Gauss-Seidel relaxation can be applied to generated smoothed vectors for feature embedding \cite{livne2012lean}, which can be implemented in linear time.

\begin{figure}\centering
\includegraphics[scale=0.3]{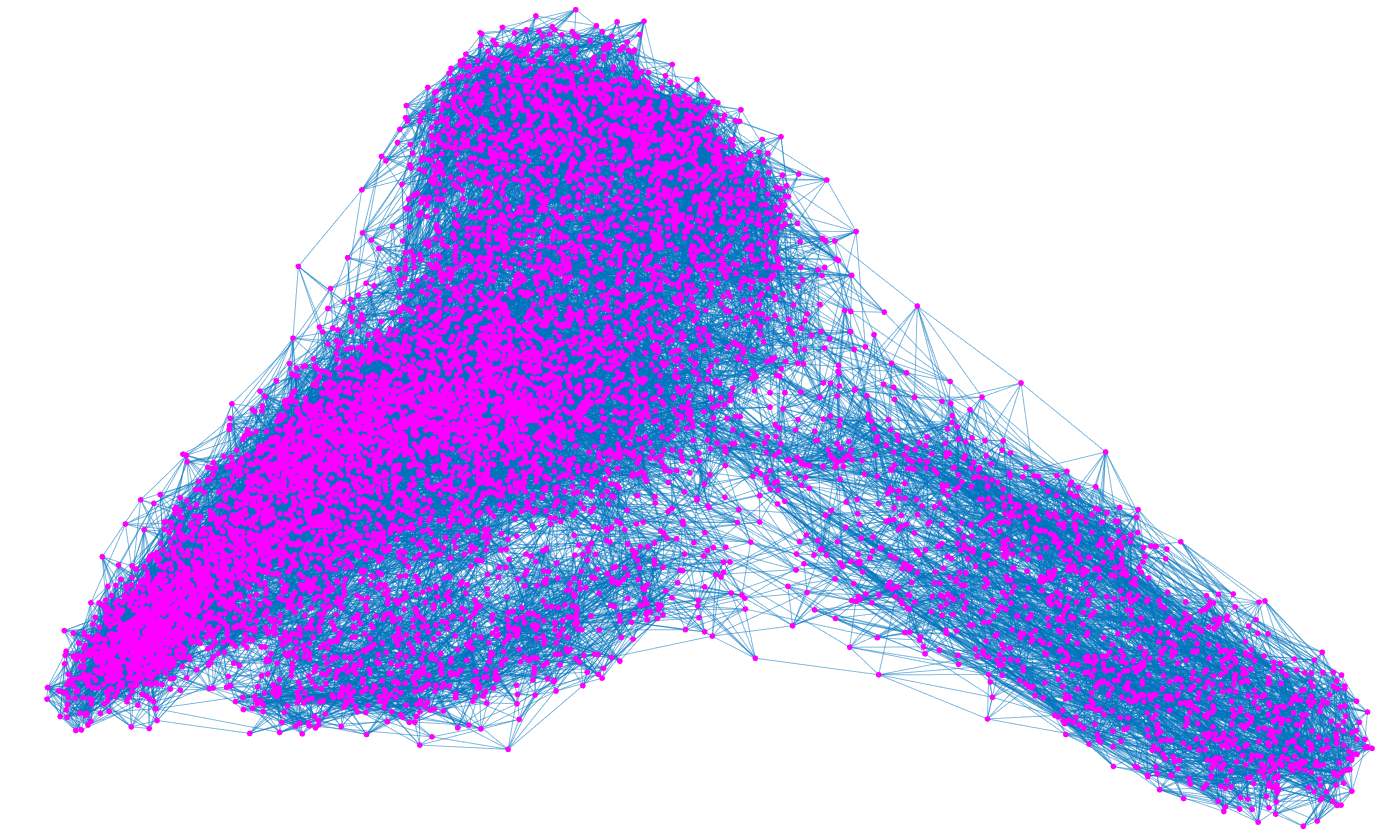}
\caption{The $k$-NN graph corresponding to the original data set (USPS).\protect\label{fig:USPS7NN}}
\end{figure}

\begin{figure}\centering
\includegraphics[scale=0.3]{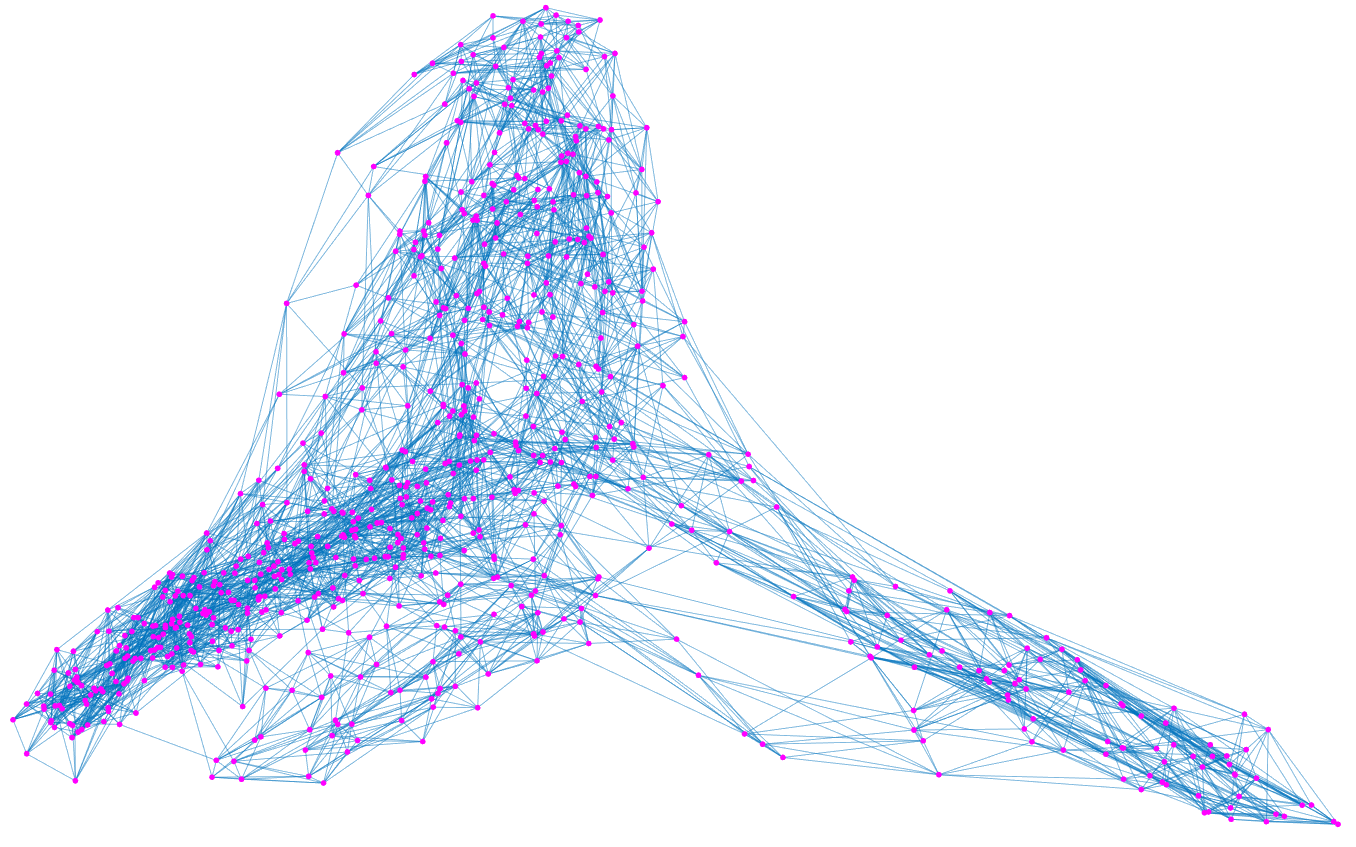}
\caption{The $k$-NN graph corresponding to the compressed data set (USPS).\protect\label{fig:USPS10X}}
\end{figure}

Based on the spectral similarity, samples are divided into different subsets. Samples in the same subset are highly correlated. We calculate the average of the feature vectors in each subset as its spectrally-representative pseudo-sample and put the representative pseudo-samples together to form the compressed data set. If the desired compression ratio is not reached, the representative pseudo-samples can be further compressed. Fig.~\ref{fig:USPS7NN} and Fig.~\ref{fig:USPS10X} show the manifolds of the original USPS data set and its compressed data set, respectively, by visualizing their corresponding $k$-NN graphs in 2D space. It can be seen that the compressed data leads to a good approximation of the manifold of the original data set.

Then, the SEP searching process can be performed on the compressed data set to efficiently find the SEPs and the downstream cluster labeling step can be applied on SEPs to assign them into different clusters. Finally, the cluster-membership of each SEP point is assigned to all of its corresponding original data points to finish the clustering task. The complete algorithm flow is shown in  Algorithm \ref{alg:kmeans}.

\textbf{Input:} A data set $D$ with $N$ samples $x_1,...,x_N \in {R}^{d}$, number of clusters $k$.\\
\textbf{Output:} Clusters $C_1$,...,$C_k$.\\

\begin{algorithmic}
    \STATE Construct a $k$-nearest neighbor ($k$-NN) graph $G$ from the input data ; \\
    \STATE Compute the Laplacian matrix corresponding to graph $G$;\\
    \STATE Perform spectrum-preserving data compression on the original data set ; \\
    \STATE Search SEP points from the compressed data set; \\
    \STATE Perform cluster-labeling on the SEP points; \\
   
    \STATE Map the cluster-memberships of SEP points back to obtain clustering result of the original data set.\\
    
\end{algorithmic}


 \subsection{Algorithm Complexity}

The spectrum-preserving data compression takes $O(|E_G|log(|V|))$ time, where  $|E_G|$ is the number of edges in the original graph and $|V|$ is the number of data points. The SEP-based SVC on the compressed data set takes $O(P^2)$ time, where $P$ is the number of compressed data points. The complexity of cluster membership retrieving is $O(N)$, where $N$ is the number of samples in the original data set.

\section{Experiment}\label{sect:experiments}

In this section, extensive experiments have been conducted to evaluate the performance of the proposed SVC method. Algorithms are performed using MATLAB running on Laptop.

\subsection{Data sets}

We use two large-scale real-world data sets to evaluate the methods: \\

\textbf{PenDigits}: A data set consists of 7,494 images of handwritten digits from 44 writers, using the sampled coordination information. Each image is represented by 16 features.\\

\textbf{USPS}: A data set with 9,298 scanned hand-written digits on the envelops from U.S. Postal Service. After processing, all the images are deslanted and size normalized, and each of them is a 16-by-16 grayscale image.

\subsection{Algorithms for Comparison}
We compare the proposed method against both the baseline and the state-of-the-art SVC methods, including:

\begin{itemize}
    \item [$\bullet$]Original SVC method (Orig SVC) \cite{ben2001support}: It Performs clustering labeling on the complete graph of the data set.
\end{itemize}

\begin{itemize}
    \item [$\bullet$]Proximity method \cite{yang2002support}: It uses proximity graph to approximate the solution of the original SVC.
\end{itemize}

\begin{itemize}
    \item [$\bullet$]SEP-based SVC \cite{lee2005improved}: It first searches SEP points and then perform complete-graph based clustering labeling only on the SEP points.
\end{itemize}

\subsection{Evaluation Metric}

The clustering quality is measured by comparing the label generated by algorithm with the ground-truth label provided by the data set. The normalized mutual information (NMI) is the most widely used metric. The NMI metric can be calculated as follows \cite{strehl2002cluster}:
\begin{equation}\label{eqn:scale}
NMI= \frac{{\sum\limits_{i = 1}^k}{\sum\limits_{j = 1}^k}{n_{i,j}}\log(\frac{{n}\cdot{n_{i,j}}}{{n_i}\cdot{n_j}}) }{{\sqrt{(\sum\limits_{i = 1}^k {{n _{i}}{\log\frac{n_i}{n}}})(\sum\limits_{j = 1}^k {{n _{j}}{\log\frac{n_j}{n}}})}}},
\end{equation}
where $n$ is the number of  data points in the data set, k is the number of clusters, $n_i$ is the number of data points in cluster $C_i$ according to the clustering result generated by algorithm, $n_j$ is the number of data points in class $C_j$ according to the ground truth labels provided by the data set, and $n_{i,j}$ is the number of data points in cluster $C_i$ according to the clustering result as well as in class $C_j$ according to the ground truth labels. The NMI value is in the range of [0, 1], while a higher NMI value indicates a better matching between the algorithm generated result and ground truth result.

\subsection{Experimental Results}

Clustering quality results of the proposed method and compared methods are provided in Table~\ref{table:nmi result}. The runtime results are reported in Table~\ref{table:runtime result}.

\begin{table}[!htbp]
\begin{center}
\scalebox{1}{
\begin{tabular}{ |c|c|c|c|c|c|c|c|c|c|c|c|c|c|c|c}
\hline Data Set&Orig SVC&Proximity&SEP-based SVC&Ours\\
\hline Pedigits  &0.56&0.59&0.79&\textbf{0.82}\\    
\hline  USPS  &0.30&0.31&0.67&\textbf{0.68}\\
\hline
\end{tabular}}
\end{center}
\caption{Clustering Quality Results (NMI)}
\label{table:nmi result}
\end{table}

\begin{table}[!htbp]
\begin{center}
\scalebox{1}{
\begin{tabular}{ |c|c|c|c|c|c|c|c|c|c|c|c|c|c|c|c}
\hline Data Set&Orig SVC&Proximity&SEP-based SVC&Ours \\
\hline Pedigits  &4308.5&357.8&1044.6&\textbf{10.0}\\    
\hline  USPS  &245.4&47.4&4665.3&\textbf{40.4}\\
\hline
\end{tabular}}
\end{center}
\caption{Runtime Results (s)}
\label{table:runtime result}
\end{table}

It can be seen from Table~\ref{table:runtime result}, for the original SEP-based SVC method, the SEP searching step and the downstream cluster labeling step take more than 1000 seconds and more than 4000 seconds for the Pendigits and USPS data sets, respectively. After spectrum-preserving data compression, only 10 second and 40 seconds are needed for the above two steps for the two data sets, respectively. Meanwhile, the proposed method achieved a even better clustering quality. Such a high performance is mainly due to the guaranteed preservation of key spectrum of the original data set. The clustering quality improvement is potentially because our spectral data compression method  allows to keep the most significant relations among data points, while avoiding noisy and misleading relationships among data points for clustering tasks. This phenomenon is also observed from the results of the Proximity method. The NMI results of the Proximity method on the two data sets are better than the results of the original SVC, mainly due to the denoising effect of the proximity graph.

As shown in  Fig. \ref{fig:nmi_svc.eps} , with increasing compression ratio, the proposed method consistently produces high clustering accuracy. This progress not only lead to dramatic speedups for SVC as shown in Fig. \ref{fig:runtime_svc.eps}, but also lead to dramatically improved memory/storage efficiency for SVC tasks. It is expected that the proposed method will be a key enabler for storing and processing much bigger data sets on more energy-efficient computing platforms, such as FPGAs  or even hand-held devices.

\begin{figure}[!h]
\centering\includegraphics[scale=0.38]{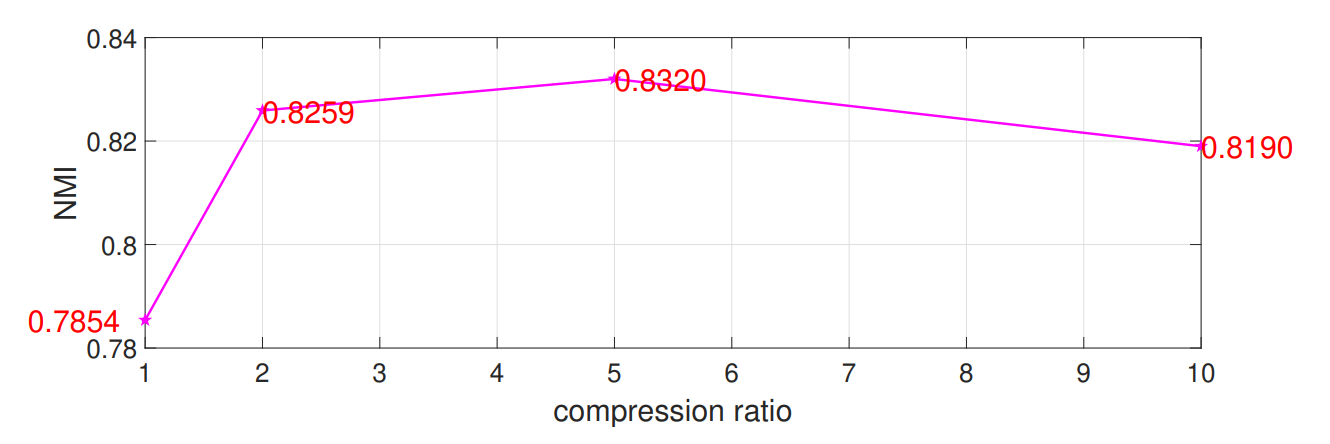}
\caption{Clustering quality VS compression ratio for the Pendigits data set.\protect\label{fig:p1.png}}
\end{figure}

\begin{figure}[!h]
\centering\includegraphics[scale=0.38]{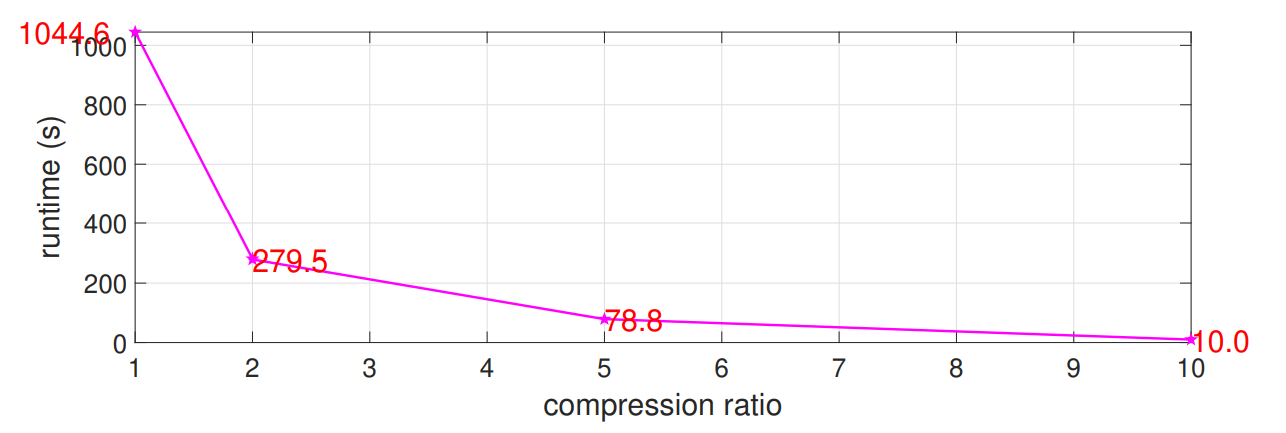}
\caption{Runtime VS compression ratio for the Pendigits data set.\protect\label{fig:p2.eps}}
\end{figure}

\section{Conclusion}\label{sect:conclusions}
To fundamentally address the computational challenge in SVC, this work applies a novel spectral data compression method to SVC that enables to construct a very small compressed data set with guaranteed preservation of the original spectrum for clustering purpose. Our results on large scale real-world data sets show dramatically improved clustering performance when compared with state-of-the-art SVC methods.

\section{Contributions}\label{sect:conclusions}
Yongyu Wang and Yuxuan Song conceived the idea. Yongyu Wang supervised, led and guided the entire project. Yongyu Wang and Yuxuan Song designed and conducted the experiments and wrote the paper. All authors discussed the results and implications and commented on the manuscript at all stages.

\nocite{*}

\bibliography{apssamp}

\begin{thebibliography}{6}%
\makeatletter
\providecommand \@ifxundefined [1]{%
 \@ifx{#1\undefined}
}%
\providecommand \@ifnum [1]{%
 \ifnum #1\expandafter \@firstoftwo
 \else \expandafter \@secondoftwo
 \fi
}%
\providecommand \@ifx [1]{%
 \ifx #1\expandafter \@firstoftwo
 \else \expandafter \@secondoftwo
 \fi
}%
\providecommand \natexlab [1]{#1}%
\providecommand \enquote  [1]{``#1''}%
\providecommand \bibnamefont  [1]{#1}%
\providecommand \bibfnamefont [1]{#1}%
\providecommand \citenamefont [1]{#1}%
\providecommand \href@noop [0]{\@secondoftwo}%
\providecommand \href [0]{\begingroup \@sanitize@url \@href}%
\providecommand \@href[1]{\@@startlink{#1}\@@href}%
\providecommand \@@href[1]{\endgroup#1\@@endlink}%
\providecommand \@sanitize@url [0]{\catcode `\\12\catcode `\$12\catcode
  `\&12\catcode `\#12\catcode `\^12\catcode `\_12\catcode `\%12\relax}%
\providecommand \@@startlink[1]{}%
\providecommand \@@endlink[0]{}%
\providecommand \url  [0]{\begingroup\@sanitize@url \@url }%
\providecommand \@url [1]{\endgroup\@href {#1}{\urlprefix }}%
\providecommand \urlprefix  [0]{URL }%
\providecommand \Eprint [0]{\href }%
\providecommand \doibase [0]{https://doi.org/}%
\providecommand \selectlanguage [0]{\@gobble}%
\providecommand \bibinfo  [0]{\@secondoftwo}%
\providecommand \bibfield  [0]{\@secondoftwo}%
\providecommand \translation [1]{[#1]}%
\providecommand \BibitemOpen [0]{}%
\providecommand \bibitemStop [0]{}%
\providecommand \bibitemNoStop [0]{.\EOS\space}%
\providecommand \EOS [0]{\spacefactor3000\relax}%
\providecommand \BibitemShut  [1]{\csname bibitem#1\endcsname}%
\let\auto@bib@innerbib\@empty
\bibitem [{\citenamefont {Ben-Hur}\ \emph {et~al.}(2001)\citenamefont
  {Ben-Hur}, \citenamefont {Horn}, \citenamefont {Siegelmann},\ and\
  \citenamefont {Vapnik}}]{ben2001support}%
  \BibitemOpen
  \bibfield  {author} {\bibinfo {author} {\bibfnamefont {A.}~\bibnamefont
  {Ben-Hur}}, \bibinfo {author} {\bibfnamefont {D.}~\bibnamefont {Horn}},
  \bibinfo {author} {\bibfnamefont {H.~T.}\ \bibnamefont {Siegelmann}},\ and\
  \bibinfo {author} {\bibfnamefont {V.}~\bibnamefont {Vapnik}},\ }\bibfield
  {title} {\bibinfo {title} {Support vector clustering},\ }\href@noop {}
  {\bibfield  {journal} {\bibinfo  {journal} {Journal of machine learning
  research}\ }\textbf {\bibinfo {volume} {2}},\ \bibinfo {pages} {125}
  (\bibinfo {year} {2001})}\BibitemShut {NoStop}%
\bibitem [{\citenamefont {Yang}\ \emph {et~al.}(2002)\citenamefont {Yang},
  \citenamefont {Estivill-Castro},\ and\ \citenamefont
  {Chalup}}]{yang2002support}%
  \BibitemOpen
  \bibfield  {author} {\bibinfo {author} {\bibfnamefont {J.}~\bibnamefont
  {Yang}}, \bibinfo {author} {\bibfnamefont {V.}~\bibnamefont
  {Estivill-Castro}},\ and\ \bibinfo {author} {\bibfnamefont {S.~K.}\
  \bibnamefont {Chalup}},\ }\bibfield  {title} {\bibinfo {title} {Support
  vector clustering through proximity graph modelling},\ }in\ \href@noop {}
  {\emph {\bibinfo {booktitle} {Proceedings of the 9th International Conference
  on Neural Information Processing, 2002. ICONIP'02.}}},\ Vol.~\bibinfo
  {volume} {2}\ (\bibinfo {organization} {IEEE},\ \bibinfo {year} {2002})\ pp.\
  \bibinfo {pages} {898--903}\BibitemShut {NoStop}%
\bibitem [{\citenamefont {Lee}\ and\ \citenamefont
  {Lee}(2005)}]{lee2005improved}%
  \BibitemOpen
  \bibfield  {author} {\bibinfo {author} {\bibfnamefont {J.}~\bibnamefont
  {Lee}}\ and\ \bibinfo {author} {\bibfnamefont {D.}~\bibnamefont {Lee}},\
  }\bibfield  {title} {\bibinfo {title} {An improved cluster labeling method
  for support vector clustering},\ }\href@noop {} {\bibfield  {journal}
  {\bibinfo  {journal} {IEEE Transactions on pattern analysis and machine
  intelligence}\ }\textbf {\bibinfo {volume} {27}},\ \bibinfo {pages} {461}
  (\bibinfo {year} {2005})}\BibitemShut {NoStop}%
\bibitem [{\citenamefont {Chung}(1997)}]{chung1997spectral}%
  \BibitemOpen
  \bibfield  {author} {\bibinfo {author} {\bibfnamefont {F.~R.}\ \bibnamefont
  {Chung}},\ }\href@noop {} {\emph {\bibinfo {title} {Spectral graph
  theory}}},\ Vol.~\bibinfo {volume} {92}\ (\bibinfo  {publisher} {American
  Mathematical Soc.},\ \bibinfo {year} {1997})\BibitemShut {NoStop}%
\bibitem [{\citenamefont {Livne}\ and\ \citenamefont
  {Brandt}(2012)}]{livne2012lean}%
  \BibitemOpen
  \bibfield  {author} {\bibinfo {author} {\bibfnamefont {O.~E.}\ \bibnamefont
  {Livne}}\ and\ \bibinfo {author} {\bibfnamefont {A.}~\bibnamefont {Brandt}},\
  }\bibfield  {title} {\bibinfo {title} {Lean algebraic multigrid (lamg): Fast
  graph laplacian linear solver},\ }\href@noop {} {\bibfield  {journal}
  {\bibinfo  {journal} {SIAM Journal on Scientific Computing}\ }\textbf
  {\bibinfo {volume} {34}},\ \bibinfo {pages} {B499} (\bibinfo {year}
  {2012})}\BibitemShut {NoStop}%
\bibitem [{\citenamefont {Strehl}\ and\ \citenamefont
  {Ghosh}(2002)}]{strehl2002cluster}%
  \BibitemOpen
  \bibfield  {author} {\bibinfo {author} {\bibfnamefont {A.}~\bibnamefont
  {Strehl}}\ and\ \bibinfo {author} {\bibfnamefont {J.}~\bibnamefont {Ghosh}},\
  }\bibfield  {title} {\bibinfo {title} {Cluster ensembles---a knowledge reuse
  framework for combining multiple partitions},\ }\href@noop {} {\bibfield
  {journal} {\bibinfo  {journal} {Journal of machine learning research}\
  }\textbf {\bibinfo {volume} {3}},\ \bibinfo {pages} {583} (\bibinfo {year}
  {2002})}\BibitemShut {NoStop}%
\end{thebibliography}%

\end{document}